%% file: main.tex
\documentclass[letterpaper,10pt,conference]{ieeeconf}
\IEEEoverridecommandlockouts
\usepackage[T1]{fontenc}
\usepackage{mathptmx}
\usepackage{amsmath,amssymb,graphicx,booktabs,xcolor,url,placeins}
\usepackage{cite}

\title{\LARGE\bf DexJoCo-X: Benchmarking Action Representations\\for Multi-Hand Dexterous Manipulation}
\author{Xiangwei Jiang$^{1}$, Yao Mu$^{2}$, Lixin Duan$^{1}$, and Wen Li$^{1,*}$%
\thanks{$^{1}$University of Electronic Science and Technology of China, China.}%
\thanks{$^{2}$Shanghai Jiao Tong University, China.}%
\thanks{$^{*}$Corresponding author: Wen Li.}}
\input{figures/fig1_firstpage.tex}
\begin{document}
\maketitle
\thispagestyle{empty}
\pagestyle{empty}
\begin{abstract}
As dexterous hands proliferate, collecting data and training policies separately for every morphology becomes increasingly impractical. Scalable cross-embodiment learning therefore requires a unified representation that captures shared manipulation structure while preserving morphology-specific control. Differences in hands, tasks, datasets, and control interfaces prevent existing studies from isolating the effects of representation, pretraining, and architecture. We introduce DexJoCo-X, a benchmark and toolkit for controlled comparison across seven representative dexterous hands, six single-arm and bimanual tasks, and 2,100 balanced demonstrations. DexJoCo-X provides a matched multi-hand, multi-task protocol with common scenes, success criteria, and execution interfaces, redesigned glove-to-hand mappings, and an automated pipeline that expands reviewed demonstrations across randomized scenes. Using $\pi_{0.5}$, Ego-Pi, and Being-H0.5, we examine whether a shared action interface is sufficient for multi-hand learning. Expanding $\pi_{0.5}$ to an 80-dimensional bimanual output yields near-zero success. Ego-Pi preserves the pretrained action head through interleaved prediction and supports per-hand multi-task learning, but remains ineffective for seven-hand joint training. By contrast, Being-H0.5 combines cross-embodiment pretraining, a unified action space, and embodiment-aware experts, enabling one policy to control all seven hands. Within this architecture, function-aligned action slots achieve 47.7\% mean success, compared with 47.0\% for native coordinates and 33.1\% for DexLatent. These results show that cross-embodiment representation depends on the entire learning system: action coordinates, pretraining, and architecture must jointly separate shared manipulation structure from embodiment-specific control.
\end{abstract}

\section{Introduction}
Learning manipulation skills that can be shared across dexterous hands is essential for deploying general-purpose robots on diverse hardware. Diverse hand kinematics and actuation fragment demonstrations and policies, making separate training costly and limiting experience reuse. Multi-embodiment datasets and generalist policies, including Open X-Embodiment/RT-X, Octo, OpenVLA, HPT, and the $\pi$ series~\cite{oxe,octo,openvla,hpt,pi0,pi05}, establish a foundation for sharing robot experience. RT-2~\cite{rt2} transfers vision--language knowledge to control. RDT-1B~\cite{rdt} uses unified action coordinates, and X-VLA~\cite{xvla} addresses embodiment heterogeneity through soft prompts, while Being-H0.5~\cite{being} and Ego-Pi~\cite{egopi} connect human demonstrations with robot learning. Complementary approaches include diffusion policies with shared action latents~\cite{lad,opfa} and cross-hand reinforcement learning~\cite{crossdex}. These efforts motivate action representations that link shared skills to embodiment-specific execution. Recent dexterous foundation-model frameworks, including UniDex~\cite{unidex2026} and XL-VLA~\cite{xlvla2026}, report promising multi-hand learning capabilities through different representations. Yet results obtained with different hands, datasets, and task protocols do not establish which approach best supports cross-hand manipulation under a common evaluation standard.

Existing benchmarks provide complementary capabilities: DexArt~\cite{dexart} studies articulated objects, Bi-DexHands~\cite{bidex} emphasizes bimanual control, DexJoCo~\cite{dexjoco2026} supplies functional tasks and demonstration tools, and DexVerse~\cite{dexverse2026} broadens task and embodiment coverage. These benchmarks expand task coverage, but leave open which action representations best support cross-hand learning under matched conditions. A fair comparison requires the same hands, tasks, demonstrations, policy backbone, and success criteria. A shared control interface is also needed to evaluate new representations without rebuilding the hand-control pipeline. These needs motivate a dedicated benchmark for fair and accessible cross-hand evaluation.

To address these needs, we introduce \textbf{DexJoCo-X}, a unified benchmark and toolkit designed for fair comparison and straightforward integration of cross-hand policies (Fig.~\ref{fig:teaser}). Built on DexJoCo, it combines seven hand interfaces, six tasks, and 2,100 demonstrations with a reusable collection pipeline. We fine-tune $\pi_{0.5}$~\cite{pi05} with the Ego-Pi protocol~\cite{egopi} as a per-hand multi-task reference, training one Native-action policy for each hand. Our central cross-hand comparison instead trains one Being-H0.5 policy~\cite{being} jointly on all seven hands under Native, FAAS, or DexLatent, and evaluates every hand on independent scenes. Our contributions are:
\begin{enumerate}
\item \textbf{A unified cross-hand benchmark.} Seven hands and six tasks share evaluation scenes, success criteria, and control interfaces for fair comparison of action representations under matched training conditions.
\item \textbf{A reusable demonstration collection toolkit.} Redesigned glove-to-hand mappings for all seven hands support human teleoperation, while automated scene expansion and trajectory verification scale demonstration collection.
\item \textbf{A balanced multi-hand demonstration dataset.} We collect 2,100 demonstrations spanning six single-arm and bimanual tasks, with 50 trajectories for each of the 42 hand--task pairs.
\end{enumerate}

\section{Related Work}
\subsection{Manipulation Benchmarks and Dexterous Tasks}
RLBench~\cite{rlbench}, CALVIN~\cite{calvin}, and LIBERO~\cite{libero} benchmark arm--gripper manipulation; Robosuite~\cite{robosuite} and ManiSkill2~\cite{maniskill2} provide broader simulation infrastructure. Arm--gripper tasks do not test independent finger coordination.

Multi-finger hand benchmarks address bimanual coordination~\cite{bidex}, articulated objects~\cite{dexart}, deformable objects~\cite{dexdeform}, and trajectory tracking across hands~\cite{pgdm}. Grasp-focused resources evaluate pose generation~\cite{dexgraspnet,gagrasp} and functional or language-guided grasping~\cite{dexfuncgrasp,dexvlg}. DexJoCo~\cite{dexjoco2026} evaluates complete functional tasks, while DexVerse~\cite{dexverse2026} expands task and embodiment coverage. Building on DexJoCo, we enable fair comparison of cross-hand action representations through balanced seven-hand, six-task demonstrations, shared evaluation scenes and success criteria, and reusable control interfaces.

\input{task_sequences_figure.tex}
\subsection{Human Demonstrations and Automated Expansion}
Human-to-robot hand mapping spans joint, Cartesian, task-oriented, and reduced-dimensional formulations~\cite{mappingreview}. DexPilot~\cite{dexpilot}, Robotic Telekinesis~\cite{telekinesis}, AnyTeleop~\cite{anyteleop}, and DIME~\cite{dime} use human motion to teach dexterous control. ALOHA~\cite{aloha} and UMI~\cite{umi} broaden accessible demonstration capture, while LEAP Hand~\cite{leap} provides an affordable dexterous embodiment.

DexMV~\cite{dexmv} and ViViDex~\cite{vividex} learn from human video demonstrations. CyberDemo~\cite{cyberdemo} augments simulated human demonstrations for real-world manipulation. MimicGen~\cite{mimicgen} expands demonstrations across contexts, and DexMimicGen~\cite{dexmimicgen} extends this approach to bimanual dexterity. Our toolkit supplies native demonstrations through seven glove mappings and verified scene expansion.

\subsection{Cross-Hand Learning and Action Representations}
Recent advances span action-chunking and diffusion policies~\cite{aloha,dp,dp4,dp3,r3dp}, vision--language--action models~\cite{rt2,openvla,pi0,pi05,xvla,being,egopi}, and world-model-based control~\cite{unipi,robodreamer,crosshandworld}. Cross-hand grasping shares contact points or maps~\cite{unigrasp,gendexgrasp,cedex}, interaction geometry~\cite{drograsp,anydexgrasp}, and morphology-conditioned eigengrasps~\cite{machagrasp}. For closed-loop cross-hand control, shared keypoint commands~\cite{crosshand2024}, retargeted human-hand actions~\cite{crossdex}, morphology-aligned graphs~\cite{dexgraspzero}, and canonical hand descriptions~\cite{canonicalhand} connect common skills to different kinematics. Particle-based world models~\cite{crosshandworld} additionally share interaction dynamics.

Shared action spaces also enable learning from heterogeneous demonstrations. LAD~\cite{lad} and OPFA~\cite{opfa} learn aligned action latents; AdvDex~\cite{advdex} aligns human and robot joints. UniDex~\cite{unidex2026} uses functional slots, while XL-VLA~\cite{xlvla2026} learns cross-hand codecs. DexJoCo-X compares these representation principles under matched data, backbones, scenes, and control interfaces across seven hands.

\section{DexJoCo-X Benchmark}
\subsection{Embodiments and Task Scope}
The seven hands are XHand, Inspire, Wuji, LEAP, Sharpa Wave, LinkerHand, and Allegro. Each retains its native kinematics, actuation, coupling, and joint limits; state and actuator dimensions are recorded separately.

Figure~\ref{fig:tasks} shows three single-arm tasks---bucket lifting, nail hammering, and Hanoi---and three bimanual tasks---microwave cooking, iPad unlocking, and photography. Together they require stable grasping, directed tool use, sequential placement, and coordination between support and operation. Task success is determined by task-specific geometric, contact, and temporal criteria.

Tasks share initial scenes, objects, and success rules across hands, enabling paired comparisons of morphology and representation. Records retain successful and failed attempts.

\subsection{Common Learning Interface and Native Control}
To compare action representations without changing the underlying control problem, DexJoCo-X separates a common learning interface from each hand's native controller. At 20\,Hz, every episode synchronizes RGB observations, task instructions, proprioceptive states, and robot commands. Arm control follows a shared convention based on world-frame translation and rotation-vector increments, whereas hand states and commands retain each embodiment's native joint order, limits, actuation, and coupling. This design preserves the control structure of each hand instead of imposing an artificial joint correspondence.

Native state and execution-command layouts are $2\times31$ and $2\times28$, with masks for valid coordinates and active arms. Encoded policy outputs have representation-specific layouts and are decoded to native commands for execution. Up to five $256\times256$ RGB streams provide front, side/ego, and wrist views, and each backbone comparison uses the same fixed view subset. Under this interface, Native, FAAS, and DexLatent share observations, arm commands, timing, and execution conditions, differing only in how hand actions are represented. Their performance can therefore be attributed more directly to the action representation and its interaction with the learning architecture.

\subsection{Training Data and Coverage}
The dataset contains 2,100 accepted demonstrations across seven hands and six tasks, with 50 trajectories per hand--task pair. All 2,100 demonstrations are used for policy training: each per-hand Ego-Pi policy receives 300 trajectories, while each joint Being-H0.5 policy receives the complete 2,100-trajectory set. Normalization statistics are computed from the corresponding training demonstrations. Policy performance is measured through separate closed-loop evaluation rollouts.

\section{Demonstration Collection}
\input{collection_figure.tex}
\subsection{Seven-Hand Teleoperation}
The capture setup follows DexJoCo~\cite{dexjoco2026}: Rokoko gloves measure finger pose, and Vive trackers measure hand motion for arm control (Fig.~\ref{fig:collection}). Each tracker attaches to the dorsal glove module through a printed connector. The operator observes the simulated task while the system records native robot states and commands.

We redesign the glove-to-hand mappings for all seven embodiments, including Allegro. Each mapping respects digit correspondence, coordinate order, motion direction, actuator limits, and joint coupling. Four-finger and five-finger hands use their respective digit configurations; dependent joints follow the native coupling model. The capture interface is
\begin{equation}
 u^{\mathrm{hand}}_k=R_h(g_k),\qquad u^{\mathrm{arm}}_k=C_h(T_k),
\end{equation}
where $h$ indexes the hand embodiment and $k$ the control step; $g_k$ is the glove measurement, $T_k\in\mathrm{SE}(3)$ is the tracked wrist pose, and $R_h$ and $C_h$ are the calibrated hand and arm mappings. The outputs $u^{\mathrm{hand}}_k$ and $u^{\mathrm{arm}}_k$ are the native hand and arm commands. These mappings generate native demonstrations. Policy-action encoding is applied afterward, so all representations share the recorded trajectories.

\subsection{Reviewed Sources and Scene Expansion}
\input{demonstration_expansion_figure.tex}
Figure~\ref{fig:demonstration_expansion} details the expansion workflow. A reviewed source pool provides candidate motions for each task. Small development trials identify failures in grasp acquisition, stage timing, tool alignment, and bimanual coordination. The resulting task-specific recipe is frozen before batch generation.

Each recipe combines scene-relative motions, stage transitions, and hand-specific parameters. Hammering varies hammer placement and nail position on the board; bimanual recipes coordinate support and operation. Attempts retain task, hand, scene seed, source, and recipe identifiers. Retries use approved sources under a fixed budget without changing the scene or success predicate.

\subsection{Verification and Export}
A candidate trajectory $\tau$ is accepted through
\begin{equation}
 A(\tau)=S(\tau)\land Q(\tau)\land R(\tau)\land D(\tau),
\end{equation}
where $A(\tau)\in\{0,1\}$ is the acceptance decision for candidate trajectory $\tau$. The binary predicates $S$, $Q$, $R$, and $D$ respectively check physical task success, motion quality, action replay, and data validity; $\land$ requires all four checks to pass. Motion checks encode task-specific requirements such as upright tool use during hammering. Replay verifies that saved commands reproduce the recorded trajectory. Export checks validate image availability, masks, timestamps, and metadata. Failed attempts and rejected exports remain in the audit record, and approved releases retain their source and configuration identifiers.

\input{action_representations.tex}

\section{Experiments}
\input{joint_training_results.tex}
\subsection{Training and Evaluation}
We evaluate two multi-task training protocols. The $\pi_{0.5}$~\cite{pi05} (Ego-Pi~\cite{egopi}) setting trains seven independent Native-action policies, one per hand, with each policy using the 300 demonstrations for that hand's six tasks. Being-H0.5~\cite{being} trains one shared policy on all 2,100 demonstrations across seven hands and six tasks; separate runs use Native, FAAS, or DexLatent actions. Table~\ref{tab:training_settings} summarizes both settings. The three Being-H0.5 runs share demonstrations, image views, arm commands, sampling, and optimization budgets. Normalization uses valid training coordinates, and the DexLatent codec remains frozen during policy training.
\input{training_settings.tex}

The $\pi_{0.5}$ (Ego-Pi) setting therefore yields seven per-hand policies, each covering six tasks, whereas each Being-H0.5 representation yields one policy covering all 42 hand--task pairs. For a given hand, both protocols use the same 50 demonstrations per task. Training samples are drawn uniformly over the applicable frames; longer trajectories therefore contribute more frames despite balanced demonstration counts.

Each hand--task cell reports the mean of three independent 50-reset evaluations, giving 150 rollouts per cell and 6,300 per seven-hand configuration. Methods share reset sets disjoint from demonstration-generation seeds. Ego-Pi outputs 50 interleaved tokens for 25 control steps, whereas Being-H0.5 outputs 30-step chunks. Decoded native commands execute at 20\,Hz with timestamped asynchronous replanning.

\subsection{Success Metrics}
For method $m$, hand $h$, task $t$, evaluation run $r$, and reset $i$, let $s_{m,h,t,r,i}\in\{0,1\}$ denote task success. We compute the hand--task success rate $\hat p_{m,h,t}$ and full-grid macro-average $\bar p_m$ as
\begin{equation}
 \hat p_{m,h,t}=\frac{1}{150}\sum_{r=1}^{3}\sum_{i=1}^{50}s_{m,h,t,r,i},\qquad
 \bar p_m=\frac{1}{42}\sum_{h=1}^{7}\sum_{t=1}^{6}\hat p_{m,h,t}.
\end{equation}
Every hand--task pair is weighted equally. Success requires complete task execution, and each single-arm or bimanual score averages 21 pairs.

\subsection{Action-Space Adaptation and Multi-Hand Results}
Before adopting interleaved prediction, we tested a direct bimanual adaptation of $\pi_{0.5}$ that reserved 40 action dimensions per side and replaced its pretrained 32-D action projection with an 80-D output head. Fine-tuning this expanded interface produced near-zero task success. We therefore follow Ego-Pi~\cite{egopi}: each native arm--hand command, containing at most 28 valid values in our benchmark, is embedded in one 32-D token, and left and right commands alternate across the output sequence. This formulation preserves the pretrained action projection and supports 25 bimanual control steps with 50 tokens.

We also trained a single Ego-Pi policy by mixing demonstrations from all seven hands, but it performed substantially below the per-hand models. A common tensor layout alone therefore did not align the heterogeneous hand-control distributions. Being-H0.5 addresses this problem earlier in learning: its pretraining maps human MANO motion and data from 30 robot embodiments into semantically aligned action slots, while its Mixture-of-Flow architecture combines shared dynamics with embodiment-aware experts~\cite{being}. This pretraining and routing provide a stronger basis for joint multi-hand optimization than downstream data mixing alone, so we use seven per-hand Ego-Pi policies as the $\pi_{0.5}$ reference and a single Being-H0.5 policy for the joint representation comparison.

Table~\ref{tab:main} shows comparable overall success for Native and FAAS under joint seven-hand training. FAAS performs better on bimanual tasks, whereas Native performs better on single-arm tasks: functional alignment provides task-dependent benefits. DexLatent achieves lower success in both groups. All three representations perform worse on bimanual tasks, highlighting coordination as a common challenge.

\section{Conclusion}
DexJoCo-X standardizes six-task evaluation across seven heterogeneous hands. Native and FAAS achieve comparable overall success, with advantages on single-arm and bimanual tasks, respectively. This tradeoff highlights task-dependent representation benefits, while bimanual coordination remains a shared challenge. Future work will extend evaluation to physical robots, broaden task coverage, and introduce held-out-hand zero-shot and one-shot protocols to assess transfer to unseen hands and adaptation from minimal demonstrations.

\section*{Acknowledgments}
OpenAI image generation tools assisted the schematic illustrations.
\FloatBarrier
\bibliographystyle{IEEEtran}
\begingroup
\let\originalthebibliography\thebibliography
\renewcommand{\thebibliography}[1]{\originalthebibliography{#1}\setlength{\baselineskip}{8.5pt}}
\bibliography{ref}
\endgroup
\end{document}

%% file: figures/fig1_firstpage.tex
\makeatletter
\newcommand{\firstpagefigurecaption}{\def\@captype{figure}\caption}
\makeatother
\IEEEaftertitletext{%
  \begin{minipage}{\textwidth}
    \centering
    \input{figures/fig1_overview_layout.tex}
    \firstpagefigurecaption[Overview of the seven-hand, six-task benchmark.]{\input{figures/fig1_caption.tex}}
    \label{fig:teaser}
  \end{minipage}\vspace{1\baselineskip}%
}

%% file: figures/fig1_overview_layout.tex
\includegraphics[width=\linewidth]{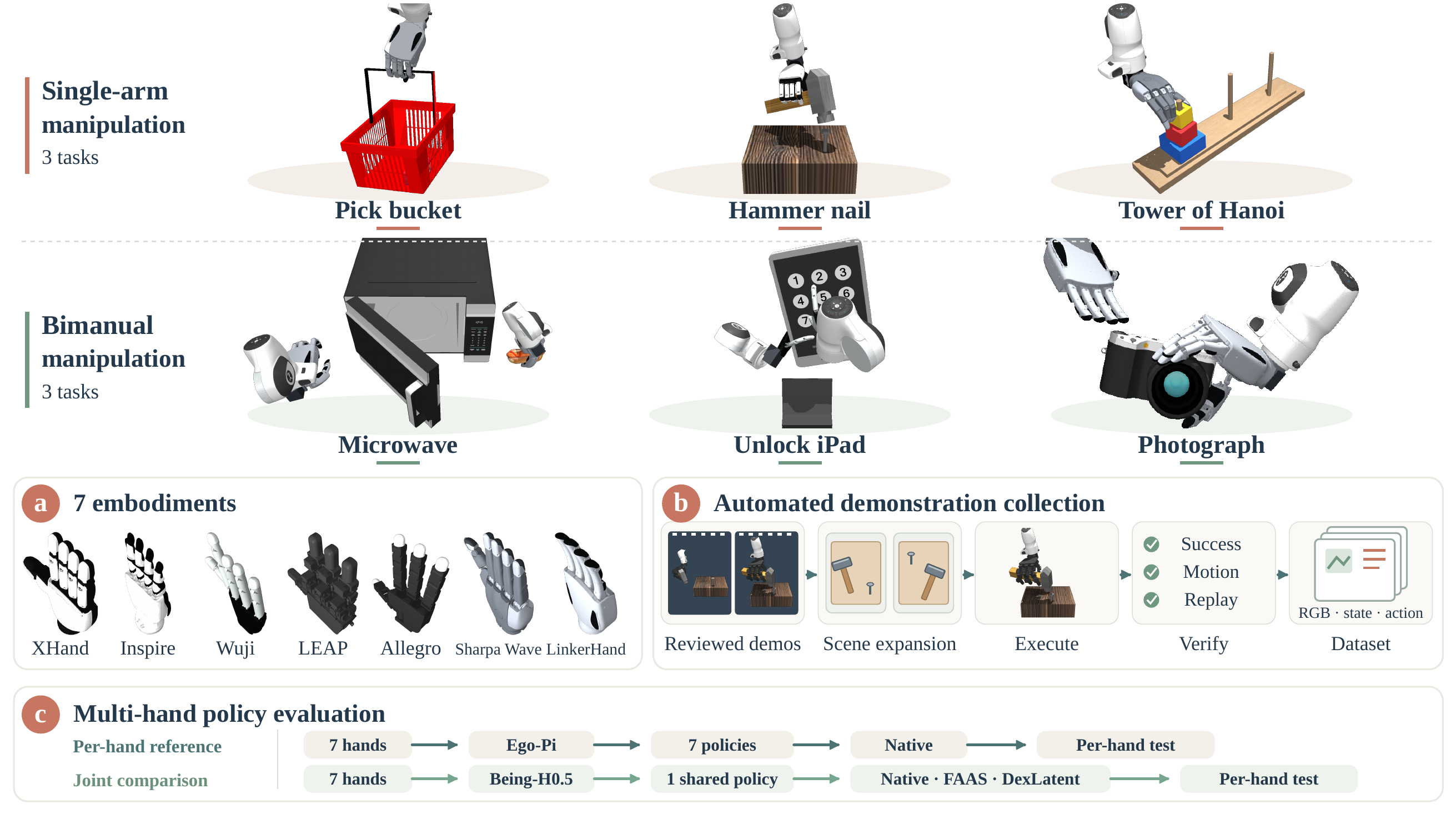}

%% file: figures/fig1_caption.tex
\textbf{DexJoCo-X overview.} Three single-arm and three bimanual tasks provide a common setting for seven dexterous hands. (a) Embodiment models. (b) Reviewed demonstrations are adapted, executed, verified, and exported with native observations and actions. (c) The evaluation includes seven per-hand $\pi_{0.5}$ (Ego-Pi) multi-task policies as a Native-action reference and one joint Being-H0.5 policy trained with Native, FAAS, or DexLatent actions; every setting is tested on all applicable hand--task pairs.

%% file: task_sequences_figure.tex
\begin{figure*}[t]
\centering
\includegraphics[width=\textwidth]{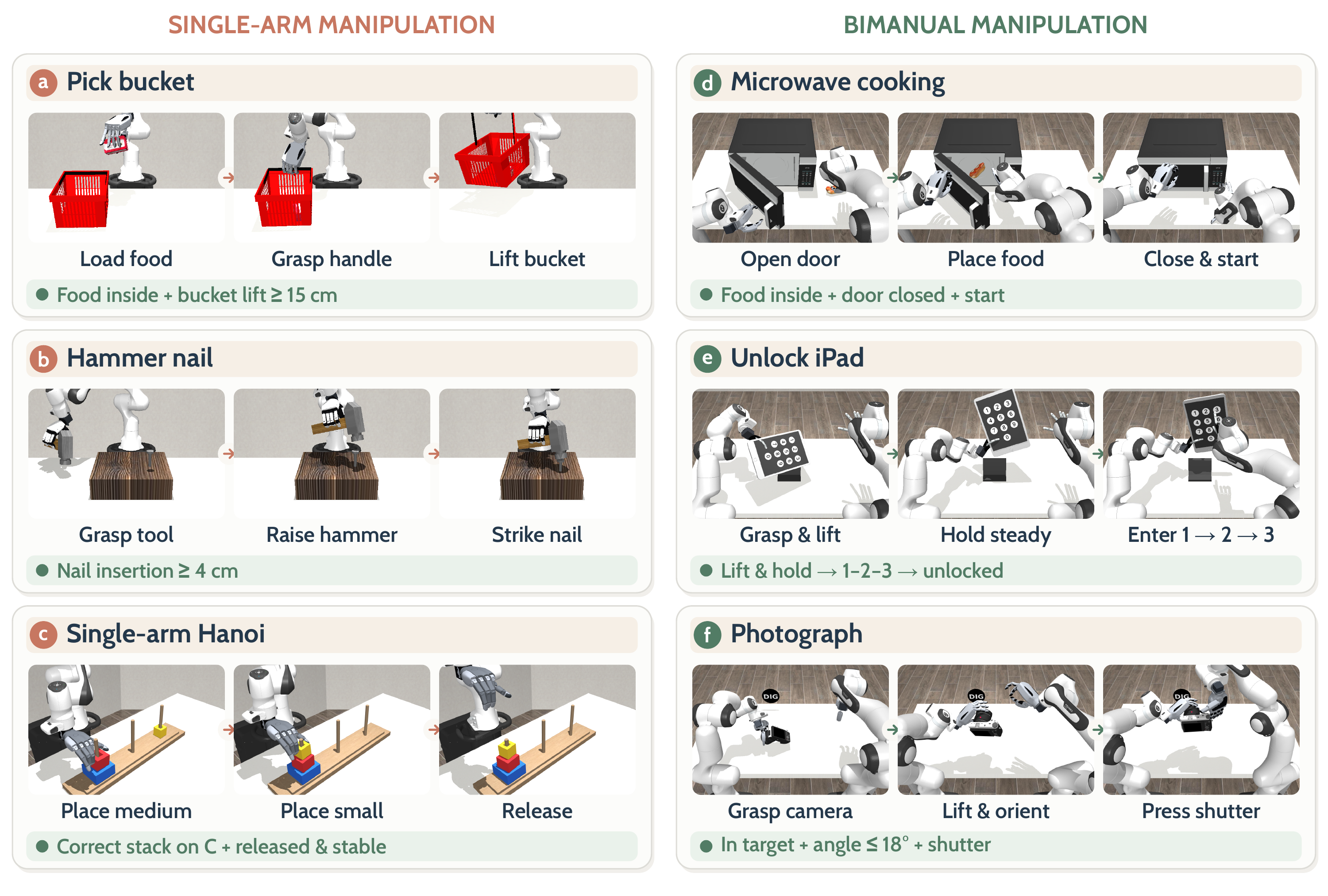}
\caption{Task stages and success conditions: (a--c) single-arm tasks; (d--f) bimanual tasks. Each sequence shows three representative stages; green labels summarize success criteria.}
\label{fig:tasks}
\end{figure*}

%% file: collection_figure.tex
\begin{figure*}[t]
\centering
\includegraphics[width=\textwidth]{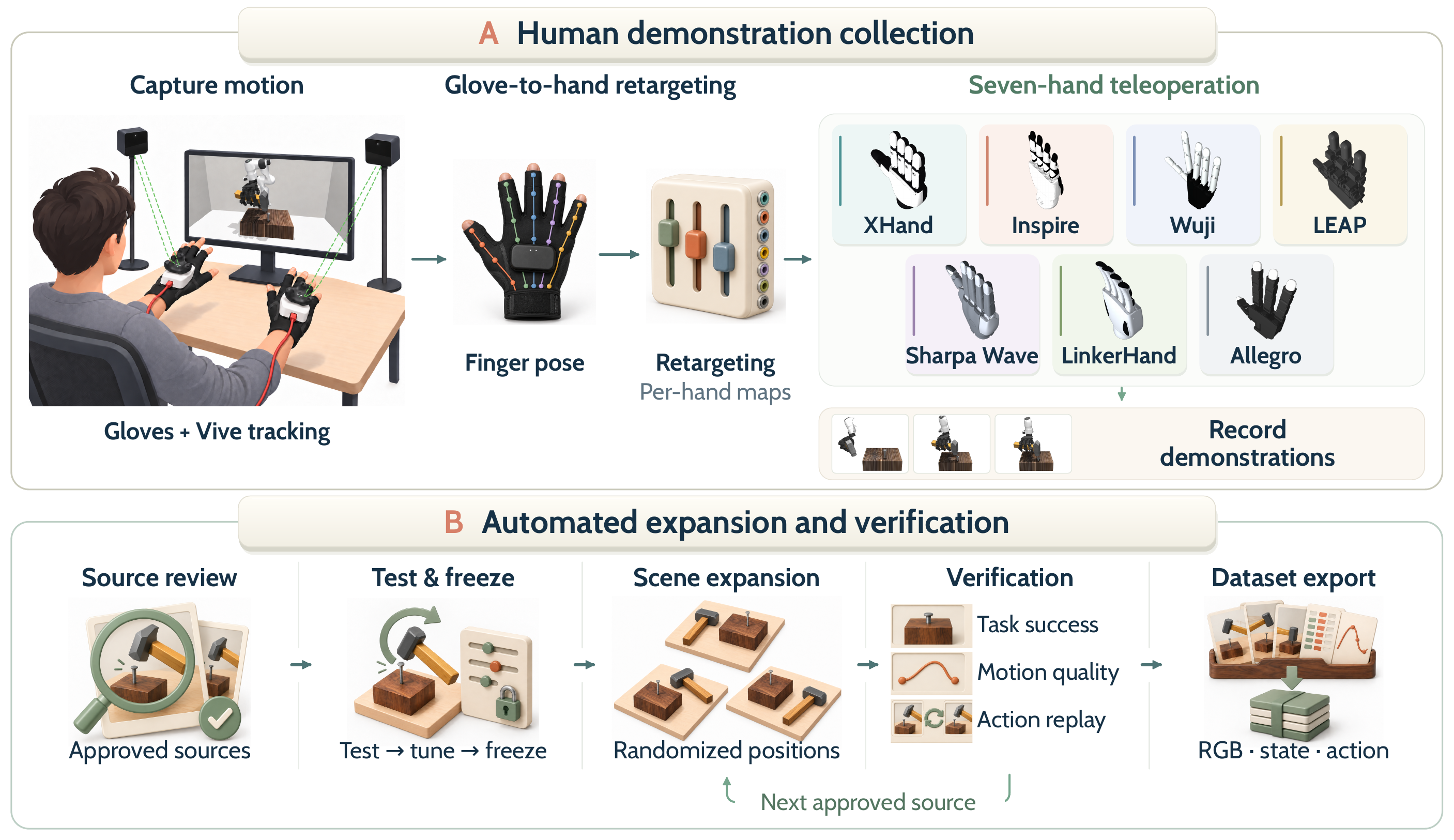}
\input{figures/collection_pipeline_caption.tex}
\end{figure*}

%% file: figures/collection_pipeline_caption.tex
\caption{Collection toolkit. (A) Glove-to-hand mappings enable teleoperation of seven hands using the DexJoCo capture setup~\cite{dexjoco2026}. (B) Reviewed demonstrations are expanded into randomized scenes, verified, and exported as RGB, state, and action records.}
\label{fig:collection}

%% file: demonstration_expansion_figure.tex
\begin{figure}[!t]
\centering
\includegraphics[width=\columnwidth]{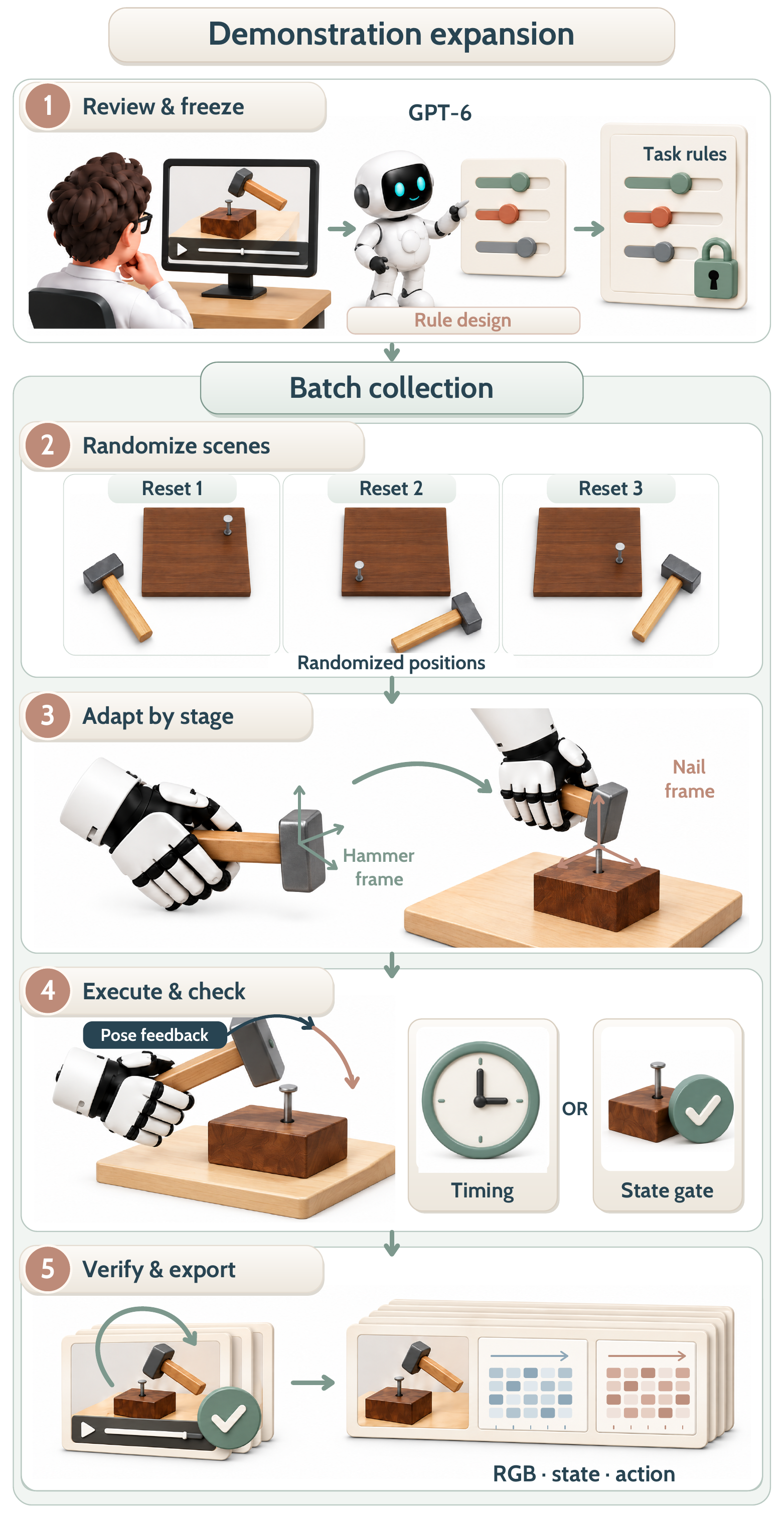}
\caption{Demonstration expansion illustrated with hammering. GPT-6 assists task-rule development; frozen rules drive batch collection. Randomized scenes are handled through stage-dependent reference frames, pose feedback, and timing- or state-based transitions. Verified trajectories are exported as RGB, state, and action records.}
\label{fig:demonstration_expansion}
\end{figure}

%% file: action_representations.tex
\section{Hand-Action Representations}
\begin{figure*}[t]
\centering
\includegraphics[width=\textwidth]{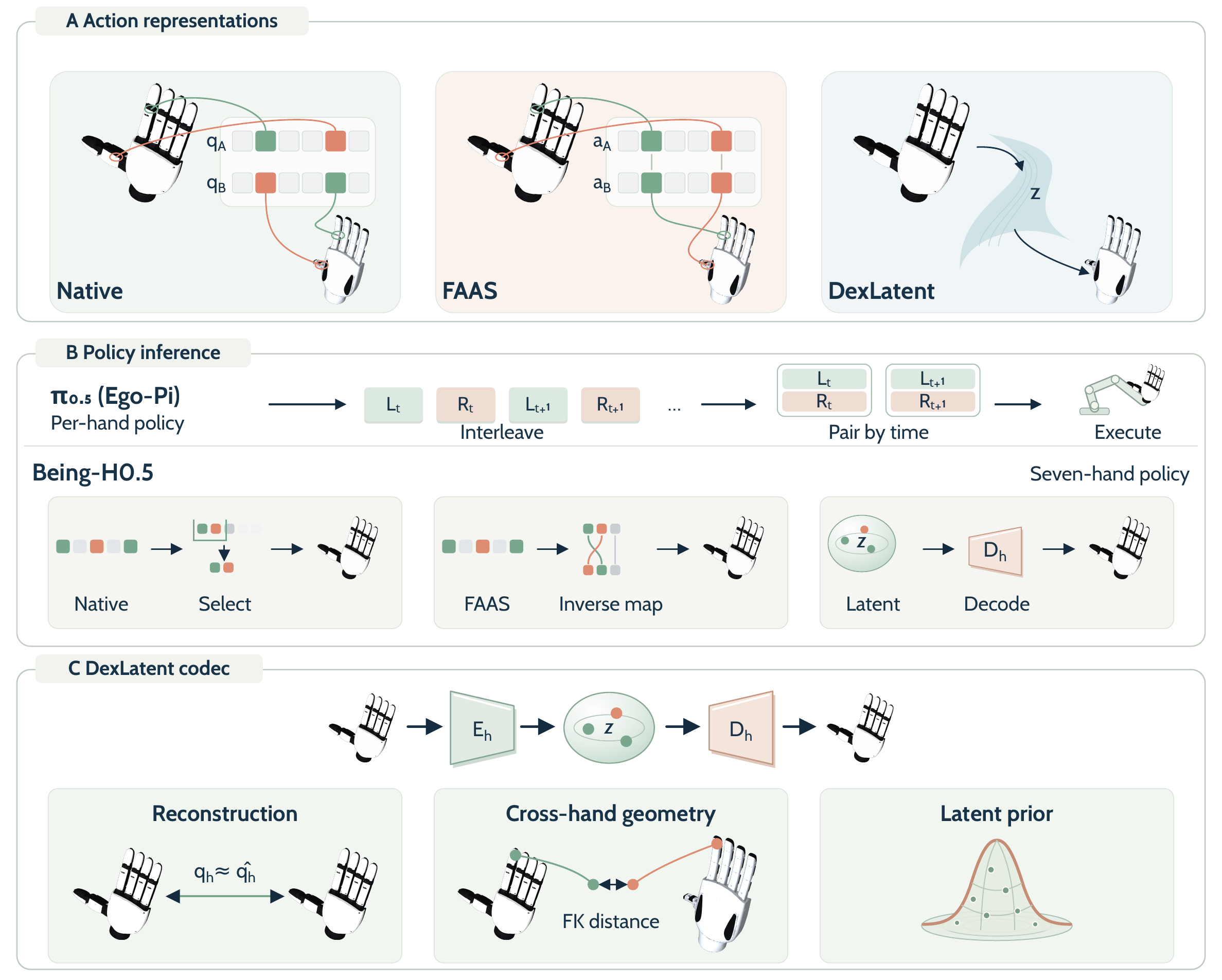}
\caption[Hand-action representations and their execution interfaces.]{\input{figures/action_representations_caption.tex}}
\label{fig:representations}
\end{figure*}

\subsection{A Common Native Execution Interface}
Let $u_h\in\mathbb R^{n_h}$ denote the valid native hand-command vector for hand $h$, where $n_h$ is its dimension. A representation supplies an encoder $e_h$ and execution decoder $d_h$: the policy predicts $z$, and the environment receives $\hat u_h=d_h(z)$. Arm commands retain the same interface across all three alternatives. Figure~\ref{fig:representations} summarizes both the coordinate meanings and the two policy interfaces. The $\pi_{0.5}$ (Ego-Pi) reference uses Native actions and interleaves left and right commands across consecutive 32-D tokens; Being-H0.5 predicts Native, FAAS, or DexLatent actions with one joint seven-hand policy.

\subsection{Native Coordinates}
Native preserves the original coordinate order of each hand. An insertion operator $P_h$ pads the valid commands to a shared dimensionality, giving $z=P_hu_h$; the decoder selects the corresponding valid coordinates. A binary mask excludes unused entries. Coordinates preserve their native meanings; the representation shares tensor dimensions across hands.

\subsection{Function-Aligned Slots}
FAAS~\cite{unidex2026} organizes commands by functional role. We encode hand commands in its 32-slot layout, separately from arm commands. These slots belong to the policy representation; decoding restores the native hand-command vector for execution. For a valid native coordinate $i$, the adapter takes the form
\begin{equation}
 z_{\sigma_h(i)}=s_{h,i}u_{h,i}+b_{h,i},
\end{equation}
where $\sigma_h$ is the functional-slot assignment and $s_{h,i}$ and $b_{h,i}$ specify direction and offset. Execution uses the corresponding inverse for active coordinates; any dependent-joint expansion must follow the hand model. The adapter aligns functional roles while retaining hand-specific command values.

\subsection{Learned Cross-Hand Latents}
DexLatent follows the hand-specific codec formulation of XL-VLA~\cite{xlvla2026}. An encoder $E_h$ maps native hand coordinates to a shared latent, and a decoder $D_h$ converts predicted latents to commands for hand $h$. Its fitting objective combines
\begin{equation}
 \mathcal L_{\mathrm{codec}}=\lambda_r\mathcal L_{\mathrm{rec}}
 +\lambda_g\mathcal L_{\mathrm{geom}}
 +\lambda_p\mathcal L_{\mathrm{prior}}.
\end{equation}
Here, $\mathcal L_{\mathrm{rec}}$, $\mathcal L_{\mathrm{geom}}$, and $\mathcal L_{\mathrm{prior}}$ respectively measure native-command reconstruction, cross-hand fingertip-geometry alignment through differentiable forward kinematics, and latent-distribution regularization; $\lambda_r$, $\lambda_g$, and $\lambda_p$ are their nonnegative weights. Geometry is evaluated on corresponding available digits. Codec checkpoints, fitting data, and calibration settings accompany each evaluated policy.

%% file: figures/action_representations_caption.tex
\textbf{Hand-action representations and policy interfaces.} Top: Native coordinates, function-aligned slots (FAAS)~\cite{unidex2026}, and DexLatent~\cite{xlvla2026}. Middle (bimanual token example): $\pi_{0.5}$ (Ego-Pi) interleaves left/right Native actions across 32-D tokens, while joint Being-H0.5 predicts and decodes the selected representation. Bottom: DexLatent fitting uses reconstruction, cross-hand geometry, and latent-prior regularization.

%% file: joint_training_results.tex
\begin{table*}[t]
\centering
\caption{Seven-hand multi-task success rates (\%), averaged over three independent 50-rollout evaluations per hand--task pair. Task entries are rounded to integers; Avg. is computed from the unrounded values over 21 pairs per panel.}
\label{tab:main}
\begingroup
\small
\setlength{\tabcolsep}{2.6pt}
\renewcommand{\arraystretch}{1.05}
\textbf{(A) Single-arm tasks}\quad Order: Bucket / Hammer / Hanoi\quad (\%)\par\vspace{3pt}
\begin{tabular*}{\textwidth}{@{\extracolsep{\fill}}llcccccccc@{}}
\toprule
Policy (training) & Representation & XHand & Inspire & Wuji & LEAP & Sharpa Wave & LinkerHand & Allegro & Avg. \\
\midrule
$\pi_{0.5}$ + Ego-Pi (per hand) & Native & 62/69/20 & 29/12/31 & 53/47/3 & 20/25/21 & 37/31/11 & 66/29/4 & 66/16/8 & 31.5 \\
\midrule
Being-H0.5 (joint) & Native & 87/46/1 & 99/74/72 & 77/85/22 & 50/39/56 & 76/63/6 & 86/58/62 & 67/35/41 & 57.2 \\
 & FAAS & 93/47/7 & 97/74/69 & 89/85/22 & 65/36/17 & 70/37/0 & 93/63/38 & 53/23/64 & 54.3 \\
 & DexLatent & 97/14/1 & 93/41/59 & 15/46/0 & 61/30/0 & 87/49/0 & 94/62/17 & 64/13/9 & 40.7 \\
\bottomrule
\end{tabular*}
\par\vspace{4pt}
\textbf{(B) Bimanual tasks}\quad Order: Microwave / Unlock iPad / Photo\quad (\%)\par\vspace{3pt}
\begin{tabular*}{\textwidth}{@{\extracolsep{\fill}}llcccccccc@{}}
\toprule
Policy (training) & Representation & XHand & Inspire & Wuji & LEAP & Sharpa Wave & LinkerHand & Allegro & Avg. \\
\midrule
$\pi_{0.5}$ + Ego-Pi (per hand) & Native & 19/38/36 & 45/9/14 & 40/1/11 & 2/1/22 & 49/1/32 & 82/36/9 & 26/0/14 & 23.2 \\
\midrule
Being-H0.5 (joint) & Native & 1/31/52 & 6/29/26 & 100/10/30 & 0/21/88 & 59/18/71 & 53/49/66 & 33/0/27 & 36.7 \\
 & FAAS & 42/64/63 & 1/44/35 & 99/1/21 & 8/29/79 & 31/8/80 & 47/79/64 & 9/1/55 & 41.0 \\
 & DexLatent & 0/5/57 & 1/1/7 & 95/5/1 & 3/17/79 & 82/13/67 & 11/16/69 & 8/0/0 & 25.6 \\
\bottomrule
\end{tabular*}
\par\vspace{4pt}
\par\vspace{3pt}\footnotesize Overall Being-H0.5 average (seven hands, six tasks): Native 47.0\%, FAAS 47.7\%, DexLatent 33.1\%.
\endgroup
\end{table*}

%% file: training_settings.tex
\begin{table}[!t]
\centering
\caption{Multi-task training settings.}
\label{tab:training_settings}
\small
\setlength{\tabcolsep}{3pt}
\begin{tabular}{@{}lcc@{}}
\toprule
Setting & $\pi_{0.5}$ + Ego-Pi & Being-H0.5 \\
\midrule
Training scope & 1 hand, 6 tasks & 7 hands, 6 tasks \\
Policies trained & 7 & 1 \\
Adaptation & Full fine-tuning & Full fine-tuning \\
Updates & 5,000 & 120,000 \\
Learning rate & $5\times10^{-5}$ & $10^{-4}$ \\
Warmup updates & 5 & 2,000 \\
LR schedule & Cosine & Cosine \\
Global batch & 128 & 8 \\
Prediction horizon & 25 steps (50 tokens) & 30 steps \\
Demonstrations & 300 per policy & 2,100 \\
\bottomrule
\end{tabular}
\end{table}